\documentclass{article} 
\usepackage{nips12submit_e,times,subfigure,amsmath,graphicx,url}

\newcommand{\vect}[1]{\mathbf{#1}}
\newcommand{\vects}[1]{\boldsymbol{#1}}
\newcommand{\matr}[1]{\mathbf{#1}}
\newcommand{\vx}[0]{\vect{x}}

\newcommand{\vy}[0]{\vect{y}}
\newcommand{\vf}[0]{\vect{f}}
\newcommand{\vh}[0]{\vect{h}}

\newcommand{\vb}[0]{\vect{b}}

\newcommand{\mA}[0]{\matr{A}}
\newcommand{\mB}[0]{\matr{B}}
\newcommand{\mC}[0]{\matr{C}}
\newcommand{\mH}[0]{\matr{H}}
\newcommand{\mI}[0]{\matr{I}}

\newcommand{\TT}[0]{\vects{\theta}}
\newcommand{\veps}[0]{\vects{\epsilon}}
\newcommand{\normal}[3]{\mathcal{N}\left(#1;#2,#3\right)}

\title{Pushing Stochastic Gradient towards Second-Order Methods
-- Backpropagation Learning with Transformations in Nonlinearities}

\author{
Tommi Vatanen, Tapani Raiko, Harri Valpola\\
Department of Information and Computer Science\\
Aalto University School of Science\\
P.O.Box 15400, FI-00076, Aalto, Espoo, Finland\\
\texttt{first.last@aalto.fi} \\
\AND
Yann LeCun\\
New York University\\
715 Broadway, New York, NY 10003, USA\\
\texttt{firstname@cs.nyu.edu}
}
\nipsfinalcopy 

\begin{document}

\maketitle

\begin{abstract}
Recently, we proposed to transform the outputs of each hidden neuron in a
multi-layer perceptron network to have zero output and zero slope on average,
and use separate shortcut connections to model the linear dependencies instead.
We continue the work by firstly introducing a third transformation to normalize
the scale of the outputs of each hidden neuron, and secondly by analyzing the
connections to second order optimization methods. We show that the
transformations make a simple stochastic gradient behave closer to second-order
optimization methods and thus speed up learning.
This is shown both in theory and with experiments. The experiments on
the third transformation show that while it further increases the speed of
learning, it can also hurt performance by converging to a worse local
optimum, where both the inputs and outputs of many hidden neurons are close to
zero.
\end{abstract}
\section{Introduction}

Learning deep neural networks has become a popular topic since the
invention of unsupervised pretraining \cite{Hinton06science}. Some later works
have returned to traditional back-propagation learning in deep models and noticed that it can
also provide impressive results \cite{Krizhevsky12} given either a sophisticated
learning algorithm \cite{Martens10} or simply enough computational power
\cite{Ciresan10simple}. In this work we study back-propagation learning in deep
networks with up to five hidden layers, continuing on our earlier results in
\cite{raiko2012}.

In learning multi-layer perceptron (MLP) networks by back-propagation, there
are known transformations that speed up learning
\cite{LeCun98,Schraudolph98decomposition,Schraudolph98factors}. For
instance, inputs are recommended to be centered to zero mean (or even whitened),
and nonlinear functions are proposed to have a range from -1 to 1 rather than 0
to 1 \cite{LeCun98}. Schraudolph
\cite{Schraudolph98factors,Schraudolph98decomposition} proposed centering all
factors in the gradient to have zero mean, and further adding linear shortcut
connections that bypass the nonlinear layer. Gradient factor centering changes
the gradient as if the nonlinear activation functions had zero mean and zero
slope on average. As such, it does not change the
model itself. It is assumed that the discrepancy between the model and the
gradient is not an issue, since the errors will be easily compensated by the
linear shortcut connections in the proceeding updates.
Gradient factor centering leads to a significant speed-up in learning.

In this paper, we transform the nonlinear activation functions in the hidden
neurons such that they have on average 1) zero mean, 2) zero slope, and 3) unit
variance. Our
earlier results in \cite{raiko2012} included the first two transformations and
here we introduce the third one.
cdsaasd
We
explain the usefulness of these transformations by studying the Fisher
information matrix and the Hessian, e.g. by measuring the angle between the
traditional gradient and a second order update direction with and without the
transformations.

It is well known that second-order optimization methods such as the natural
gradient \cite{Amari98} or Newton's method decrease the number of required
iterations compared to the basic gradient descent, but they cannot be easily
used with high-dimensional models due to heavy computations with
large matrices. In practice, it is possible to use a diagonal or block-diagonal
approximation \cite{LeRoux08} of the Fisher information matrix or the Hessian.
Gradient descent can be seen as an approximation of the second-order methods,
where the matrix is approximated by a scalar constant times a unit matrix. Our
transformations aim at making the Fisher information matrix as close to such
matrix as possible, thus diminishing the difference between first and second
order methods. Matlab code for replicating the experiments in this paper is
available at
\begin{center}
 \url{https://github.com/tvatanen/ltmlp-neuralnet}
\end{center}

\section{Proposed Transformations}

Let us study a MLP-network with a single hidden layer and shortcut mapping, that
is, the output column vectors $\vy_t$ for each sample $t$ are modeled as a
function of the input column vectors $\vx_t$ with
\begin{equation}
  \vy_t = \mA \vf\left( \mB \vx_t \right) + \mC \vx_t + \veps_t,
  \label{eq:mlp_gaussian}
\end{equation}
where $\vf$ is a nonlinearity (such as $\tanh$) applied to each component of
the argument vector separately, $\mA$, $\mB$, and $\mC$ are the weight matrices,
and $\veps_t$ is the noise which is assumed to be zero mean and Gaussian, that
is, $p(\epsilon_{it})=\normal{\epsilon_{it}}{0}{\sigma_i^2}$. In order to avoid
separate bias vectors that complicate formulas, the input vectors are assumed
to have been supplemented with an additional component that is always one.

Let us supplement the $\tanh$ nonlinearity with auxiliary scalar
variables $\alpha_i$, $\beta_i$, and $\gamma_i$ for each nonlinearity $f_i$.
They are updated before each gradient evaluation in order to help learning of
the other parameters $\mA$,  $\mB$ , and $\mC$. We define
\begin{equation}
  f_i(\vb_i \vx_t) = \gamma_i \left[ \tanh(\vb_i \vx_t)+\alpha_i \vb_i \vx_t +
\beta_i \right]
  \label{eq:transformation},
\end{equation}
where $\vb_i$ is the $i$th row vector of matrix $\mB$.
We will ensure that
\begin{align}
  \sum_{t=1}^T f_i(\vb_i \vx_t) &= 0 \label{eq:zeromean} \\
  \sum_{t=1}^T f_i^\prime(\vb_i \vx_t) &= 0 \label{eq:zeroslope} \\
  \left[\sum_{t=1}^T \frac{f_i(\vb_i \vx_t)^2}{T}\right]
  \left[\sum_{t=1}^T \frac{f_i^\prime(\vb_i \vx_t)^2}{T}\right] &= 1
\label{eq:unitvar}
\end{align}
by setting $\alpha_i$, $\beta_i$, and $\gamma_i$ to
\begin{align}
\label{eq:alpha}
   \alpha_i &= -\frac{1}{T}\sum_{t=1}^T \tanh^\prime(\vb_i \vx_t)  \\
   \beta_i &= -\frac{1}{T} \sum_{t=1}^T \left[ \tanh(\vb_i \vx_t) + \alpha_i
   \vb_i\vx_t \right] \label{eq:beta} \\
   \gamma_i &= \Big\{ \frac{1}{T} \sum_{t=1}^T \left[ \tanh(\vb_i
\vx_t)+\alpha_i \vb_i \vx_t + \beta_i \right]^2\Big\}^{1/4} 
 \Big\{ \frac{1}{T} \sum_{t=1}^T \left[ \tanh^\prime(\vb_i
\vx_t)+\alpha_i \right]^2\Big\}^{1/4} \label{eq:gamma}.
\end{align}

One way to motivate the first two transformations in Equations
\eqref{eq:zeromean} and \eqref{eq:zeroslope}, is to study the expected output
$\vy_t$ and its dependency of the input $\vx_t$:
\begin{align}
   \frac{1}{T} \sum_t \vy_t &= \mA \frac{1}{T} \sum_t \vf(\mB \vx_t) + \mC
\frac{1}{T} \sum_t \vx_t \label{eq:expected_y}\\
   \frac{1}{T} \sum_t \frac{\partial \vy_t}{\partial \vx_t} &= \mA\left[
\frac{1}{T} \sum_t \vf^\prime(\mB\vx_t)\right] \mB^T + \mC.
\label{eq:expected_dydx}
\end{align}
We note that by making nonlinear activations $\vf(\cdot)$ zero mean in Eq.\ 
\eqref{eq:zeromean}, we disallow the nonlinear mapping $\mA
\vf\left( \mB \cdot \right)$ to affect the
expected output $\vy_t$, that is, to compete with the bias term. Similarly, by
making the nonlinear activations $\vf(\cdot)$ zero slope in Eq.\
\eqref{eq:zeroslope}, we disallow the nonlinear mapping $\mA \vf\left( \mB \cdot
\right)$ to affect the expected dependency of the input, that is, to compete
with the linear mapping $\mC$. In traditional neural networks, the linear
dependencies (expected $\partial \vy_t / \partial \vx_t$) are
modeled by many competing paths from an input to an output (e.g.\ via each
hidden unit), whereas our architecture gathers the linear dependencies to be
modeled only by $\mC$. We argue that less competition between parts of the model
will speed up learning. Another explanation for choosing these transformations
is that they make the nondiagonal parts of the Fisher information matrix closer
to zero (see Section~\ref{sec:fisher}).

The goal of Equation \eqref{eq:unitvar} is to normalize both
the output signals (similarly as data is often normalized as a preprocessing
step -- see,e.g., \cite{LeCun98}) and the slopes of the output signals of each
hidden unit at the same time. This is motivated by observing that the diagonal
of the Fisher information matrix contains elements with both the signals and
their slopes. By these normalizations, we aim pushing these diagonal elements
more similar to each other. As we cannot normalize both the signals and the
slopes to unity at the same time, we normalize their geometric mean to unity. 

The effect of the first two transformations can be compensated exactly by updating the shortcut mapping $\mC$ by
\begin{align}
  \mC_\text{new} = \mC_\text{old} &-
\mA(\boldsymbol{\alpha}_\text{new}-\boldsymbol{\alpha}_\text{old})\mB
\nonumber \\
  &-\mA(\vects{\beta}_\text{new}-\vects{\beta}_\text{old})
  \left[0\phantom{0}0\dots 1\right], \label{eq:shortcutupdate}
\end{align}
where $\boldsymbol{\alpha}$ is a matrix with elements $\alpha_i$ on the
diagonal and one empty row below for the bias term, and $\vects{\beta}$ is a
column vector with components $\beta_i$ and one zero below for the bias term.
The third transformation can be compensated by
\begin{align}
  \mA_\text{new} =\mA_\text{old} \boldsymbol{\gamma}_\text{old}
  \boldsymbol{\gamma}_\text{new}^{-1},
\end{align}
where $\boldsymbol{\gamma}$ is a diagonal matrix with $\gamma_i$ as the diagonal
elements.




Schraudolph \cite{Schraudolph98factors,Schraudolph98decomposition} proposed
centering the factors of the gradient to zero mean. It was argued that
deviations from the gradient fall into the linear subspace that the shortcut
connections operate in, so they do not harm the overall performance.
Transforming the nonlinearities as proposed in this paper has a similar
effect on the gradient. Equation \eqref{eq:zeromean} corresponds to
Schraudolph's \emph{activity centering} and Equation (\ref{eq:zeroslope})
corresponds to \emph{slope centering}.

\section{Theoretical Comparison to a Second-Order Method}
\label{sec:fisher}

Second-order optimization methods, such as the natural gradient \cite{Amari98}
or Newton's method, decrease the number of required iterations compared to the
basic gradient descent, but they cannot be easily used with large models due to
heavy computations with large matrices. The natural gradient is the basic
gradient multiplied from the left by the inverse of the Fisher information
matrix. Using basic gradient descent can thus be seen as using the natural
gradient while approximating the Fisher information with a unit matrix
multiplied by the inverse learning rate. We will show how the first two proposed
transformations move the non-diagonal elements of the Fisher information matrix
closer to zero, and the third transformation makes the diagonal elements more
similar in scale, thus making the basic gradient behave closer to the natural
gradient.

The Fisher information matrix contains elements
\begin{equation}
  G_{ij} = \sum_t \left< \frac{\partial^2 \log p(\vy_t \mid \vx_t, \mA, \mB,
  \mC )} {\partial \theta_i \partial \theta_j} \right>,
\end{equation}
where $\left<\cdot\right>$ is the expectation over the Gaussian
distribution of noise $\veps_t$ in Equation (\ref{eq:mlp_gaussian}), and
vector $\TT$ contains all the elements of matrices $\mA$, $\mB$, and $\mC$.
Note that here $\vy_t$ is a random variable and thus the Fisher information does
not depend on the output data. The Hessian matrix is closely related to the
Fisher information, but it does depend on the output data and contains more
terms, and therefore we show the analysis on the simpler Fisher information
matrix.

The elements in the Fisher information matrix are:
\begin{equation}
  \frac{\partial}{\partial a_{ij}}\frac{\partial}{\partial a_{i^\prime
 j^\prime}} \log p =
\left\{
  \begin{array}{ll} 0 & i^\prime\neq i \\
      - \frac{1}{\sigma_i^2} \sum_t
f_j(\vb_j\vx_t)f_{j^\prime}(\vb_{j^\prime}\vx_t) & i^\prime=i,
  \end{array} \right. \label{eq:fisher_a_a}
\end{equation}
where $a_{ij}$ is the $ij$th element of matrix $\mA$, $f_j$ is the $j$th
nonlinearity, and $\vb_j$ is the $j$th row vector of matrix $\mB$. Similarly
\begin{align}
  \frac{\partial}{\partial b_{jk}}\frac{\partial}{\partial b_{j^\prime
  k^\prime}} \log p =
  - \sum_i \frac{1}{\sigma_i^2} a_{ij} a_{ij^\prime} \sum_t
 f_j^\prime(\vb_j \vx_t) f_{j^\prime}^\prime(\vb_{j^\prime}
  \vx_t) x_{kt} x_{k^\prime t}
  \label{eq:fisher_b_b}
\end{align}
and
\begin{equation}
  \frac{\partial}{\partial c_{ik}}\frac{\partial}{\partial c_{i^\prime
   k^\prime}} \log p = \left\{
  \begin{array}{ll} 0 & i^\prime\neq i \\
      - \frac{1}{\sigma_i^2} \sum_t x_{kt} x_{k^\prime t} & i^\prime=i.
  \end{array} \right. \label{eq:fisher_c_c}
\end{equation}
The cross terms are
\begin{align}
  \frac{\partial}{\partial a_{ij}}\frac{\partial}{\partial b_{j^\prime k}} \log
   p &= - \frac{1}{\sigma_i^2} a_{ij^\prime}
    \sum_t f_j(\vb_j\vx_t) f_{j^\prime}^\prime(\vb_{j^\prime}\vx_t)
   x_{kt} \label{eq:fisher_a_b} \\
  \frac{\partial}{\partial c_{ik}}\frac{\partial}{\partial a_{i^\prime j}} \log
p &=
  \left\{ \begin{array}{ll} 0 & i^\prime \neq i \\
      - \frac{1}{\sigma_i^2} \sum_t f_j(\vb_j\vx_t) x_{kt} & i^\prime =i
  \end{array} \right. \label{eq:fisher_a_c} \\
  \frac{\partial}{\partial c_{ik}}\frac{\partial}{\partial b_{jk^\prime}} \log p
   &= - \frac{1}{\sigma_i^2} a_{ij} \sum_t f_j^\prime(\vb_j\vx_t) x_{kt}
   x_{k^\prime t}.
  \label{eq:fisher_b_c}
\end{align}

Now we can notice that Equations
(\ref{eq:fisher_a_a}--\ref{eq:fisher_b_c}) contain factors such as
$f_j(\cdot)$, $f_j^\prime(\cdot)$,
and $x_{it}$. We argue that by making the factors as close to zero as possible,
we help in making nondiagonal elements of the Fisher information closer to zero.
For instance, $E[f_j(\cdot)f_{j^\prime}(\cdot)] =
E[f_j(\cdot)]E[f_{j^\prime}(\cdot)] +
\text{Cov}[f_j(\cdot),f_{j^\prime}(\cdot)]$, so assuming that the hidden units $j$ and
$j^\prime$ are representing different things, that is, $f_j(\cdot)$ and
$f_{j^\prime}(\cdot)$
are uncorrelated, the nondiagonal element of the Fisher information in Equation
(\ref{eq:fisher_a_a}) becomes exactly zero by using the transformations. When
the units are not completely uncorrelated, the element in question will be
only approximately zero.
The same argument applies to all other elements in Equations
(\ref{eq:fisher_b_b}--\ref{eq:fisher_b_c}), some of them also highlighting the
benefit of making the input data $\vx_t$ zero-mean.
Naturally, it is unrealistic to assume that inputs $\vx_t$, nonlinear
activations $\vf(\cdot)$, and their slopes $\vf^\prime(\cdot)$ are all
uncorrelated, so the goodness of this approximation is empirically evaluated in
the next section.

The diagonal elements of the Fisher can be found in Equations
(\ref{eq:fisher_a_a}--\ref{eq:fisher_c_c}) when $i=i^\prime$, $j=j^\prime$, and
$k=k^\prime$. There we find $f(\cdot)^2$ and $f^\prime(\cdot)^2$ that we aim to
keep similar in scale by using the third transformation in Equation
\eqref{eq:unitvar}.

\section{Empirical Comparison to a Second-Order Method}
\label{sec:hessian}

Here we investigate how linear transformations affect the gradient by comparing
it to a second-order method, namely Newton's algorithm with a simple
regularization to make the Hessian invertible. 

We compute an approximation of the Hessian matrix using finite difference
method, in which case $k$-th row vector $\vh_k$ of the Hessian matrix $\mH$ is
given by
\begin{equation}\label{eq:hessian}
 \vh_k = \frac{\partial(\nabla E(\TT))}{\partial \theta_k} \approx
 \frac{\nabla E(\TT + \delta \boldsymbol{\phi}_k) - \nabla E(\TT - \delta
\boldsymbol{\phi}_k)}{2\delta},
\end{equation}
where $\boldsymbol{\phi}_k = (0,0,\dots,1,\dots,0)$ is a vector of zeros and 1
at the $k$-th position, and the error function $E(\TT)=-\sum_t \log p(\vy_t
\mid \vx_t, \TT)$. The resulting Hessian might still contain some very small or
even negative eigenvalues which cause its inversion to blow up. 
Therefore we do not use the Hessian directly, but include a regularization term
similarly as in the Levenberg-Marquardt algorithm,
resulting in a second-order update direction
\begin{equation}\label{eq:secondorder}
 \Delta \TT = (\mH + \mu \mI)^{-1} \nabla E(\TT), 
\end{equation}
where $\mI$ denotes the unit matrix. 
Basically, Equation
\eqref{eq:secondorder} combines the steepest descent and the second-order
update rule in such a way, that when $\mu$ gets small, the update direction
approaches the Newton's method and vice versa.

Computing the Hessian is computationally demanding and therefore we have to
limit the size of the network used in the experiment. We study the MNIST
handwritten digit classification problem where the dimensionality of the
input data has been reduced to 30 using PCA with a random rotation
\cite{raiko2012}. We use a network with two hidden layers with architecture
30--25--20--10. The network was trained using the standard gradient descent
with weight decay regularization. Details of the training are given in
the appendix. 

In what follows, networks with all three transformations (\emph{LTMLP},
linearly transformed multi-layer perceptron network), with two transformations
(\emph{no-gamma} where all $\gamma_i$ are fixed to unity) and a network with no
transformations (\emph{regular}, where we fix $\alpha_i=0$, $\beta_i=0$, and
$\gamma_i=1$) were compared. The Hessian matrix was approximated according to
Equation \eqref{eq:hessian} 10 times in regular intervals during the training of
networks. All figures are shown using the approximation after 4000 epochs of
training, which roughly corresponds to the midpoint of learning. However, the
results were parallel to the reported ones all along the training.

\begin{figure}[tb]
\centering
\subfigure[Eigenvalues]{
\includegraphics[width=4.9cm]{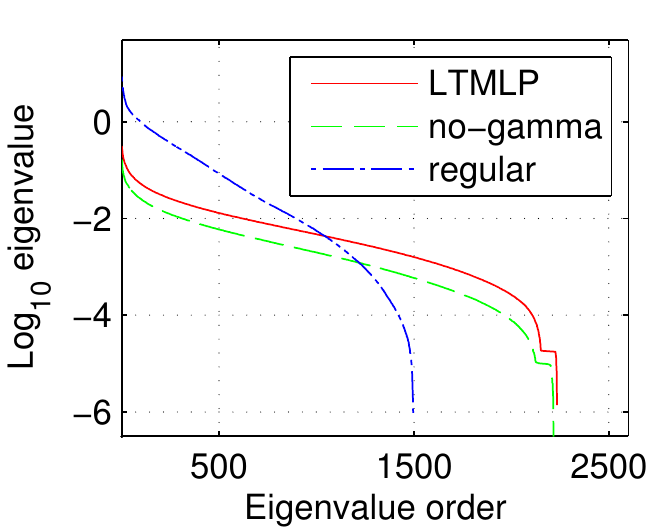}
}~
\subfigure[Angles]{
\includegraphics[width=4.9cm]{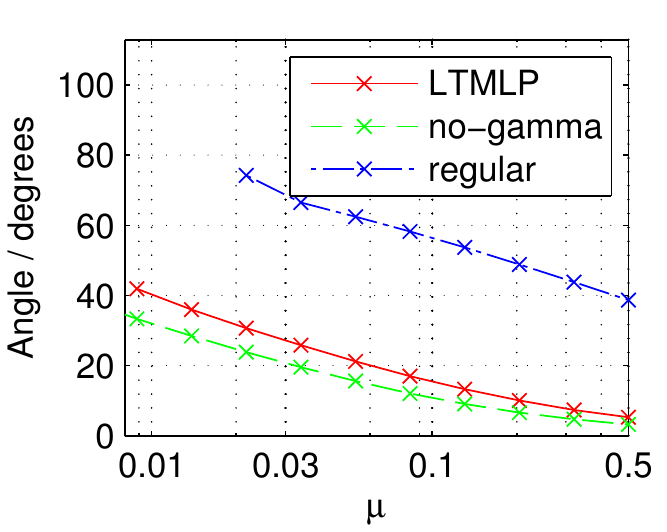}
}
\caption{Comparison of (a) distributions of the eigenvalues of Hessians
($2600\times2600$ matrix) and (b) angles compared to the second-order update
directions using LTMLP and regular MLP. In (a), the eigenvalues are distributed
most evenly when using LTMLP. (b) shows that gradients of the transformed
networks point to the directions closer to the second-order update.}
\label{fig:eigvalues}
\end{figure}

We studied the eigenvalues of the Hessian matrix ($2600\times2600$) and the
angles between the methods compared and second-order update direction.
The distribution of eigenvalues in Figure~\ref{fig:eigvalues}a for the
networks with transformations are more even compared to the regular MLP.
Furthermore, there are fewer negative eigenvalues, which are not shown in the
plot, in the transformed networks. In Figure~\ref{fig:eigvalues}b, the angles
between the gradient and the second-order update direction are compared as
a function of $\mu$ in Equation \eqref{eq:secondorder}. The plots are cut when
$\mH + \mu \mI$ ceases to be positive definite as $\mu$ decreases. Curiously,
the update directions are closer to the second-order method, when $\gamma$ is
left out, suggesting that $\gamma$s are not necessarily useful in this respect.

\begin{figure}[tb]
\centering
\subfigure[LTMLP]{
\includegraphics[width=3.9cm]{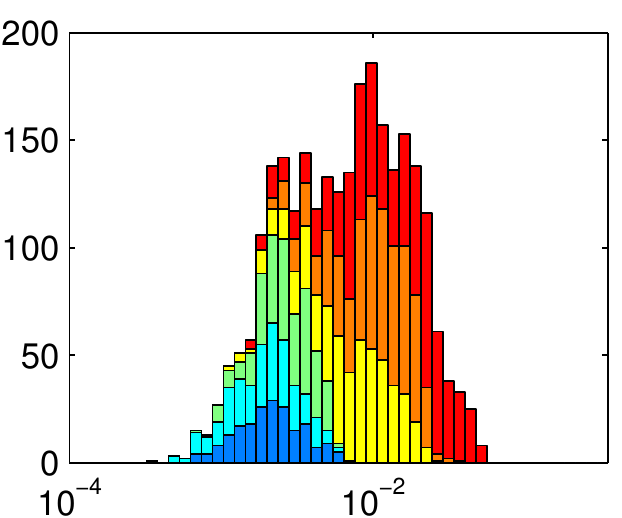}
}~
\subfigure[no-gamma]{
\includegraphics[width=3.9cm]{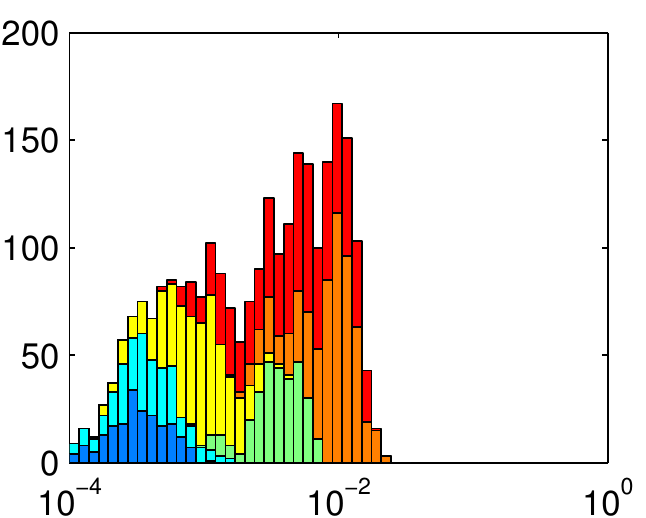}
}~
\subfigure[regular]{
\includegraphics[width=3.9cm]{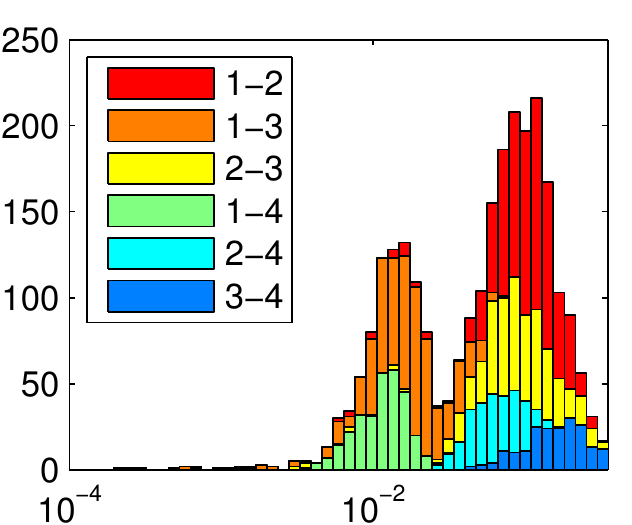}
}
\caption{Comparison of distributions of the diagonal elements of Hessians.
Coloring according to legend in (c) shows which layers to corresponding weights
connect (1 = input, 4 = output). Diagonal elements are most concentrated
in LTMLP and most spread in the regular MLP network. Notice the
logarithmic x-axis.}
\label{fig:diagvalues}
\end{figure}

Figure~\ref{fig:diagvalues} shows histograms of the diagonal elements
of the Hessian
after 4000 epochs of training. All the distributions are bimodal, but the
distributions are closer to unimodal when transformations are used (subfigures
(a) and (b))\footnote{It can be also argued whether (a) is more unimodal
compared to (b).}. Furthermore, the variance of the diagonal elements in log-scale is smaller when using 
LTMLP, $\sigma^2_\text{a}=0.90$, compared to the
other two, $\sigma^2_\text{b} = 1.71$ and $\sigma^2_\text{c} =
1.43$.
This suggests that when transformations are used, the second-order update rule
in Equation \eqref{eq:secondorder} corrects different elements of the gradient
vector more evenly compared to a regular back-propagation learning, implying
that the gradient vector is closer to the second-order update direction when
using all the transformations.

To conclude this section, there is no clear evidence in way or another whether
the addition of $\gamma$ benefits the back-propagation learning with only
$\alpha$ and $\beta$. However, there are some differences between these two
approaches. In any case, it seems clear that transforming the nonlinearities
benefits the learning compared to the standard back-propagation learning.

\section{Experiments: MNIST Classification}
\label{sec:classification}

We use the proposed transformations for training MLP networks for MNIST
classification task. Experiments are conducted without pretraining,
weight-sharing, enhancements of the training set or any other known tricks to
boost the performance. No weight decay is used and as only
regularization we add Gaussian noise with $\sigma=0.3$ to the training data.
Networks with two and three hidden layers with architechtures 784--800--800--10
(solid lines) and 784--400--400--400--10 (dashed lines) are used. Details are
given in the appendix.

Figure~\ref{fig:mnist} shows the results as number of errors in classifying the
test set of 10\ 000 samples. The results of the regular back-propagation without
transformations, shown in blue, are well in line with previously published
result for this task. 
When networks with same architecture are
trained using the proposed transformations, the results are improved
significantly. However, adding $\gamma$ in addition to previously proposed
$\alpha$ and $\beta$ does not seem to affect results on this data set. The best
results, 112 errors, is obtained by the smaller architecture without $\gamma$
and for the three-layer architecture with $\gamma$ the result is 114 errors.
The learning seems to converge faster, especially in the three-layer case, with
$\gamma$. The results are in line what was obtained in \cite{raiko2012} where
the networks were regularized more thoroughly. These results show that it is
possible to obtain results comparable to dropout networks (see
\cite{hinton2012}) using only minimal regularization.

\begin{figure}[tb]
\centering
\includegraphics[width=7cm]{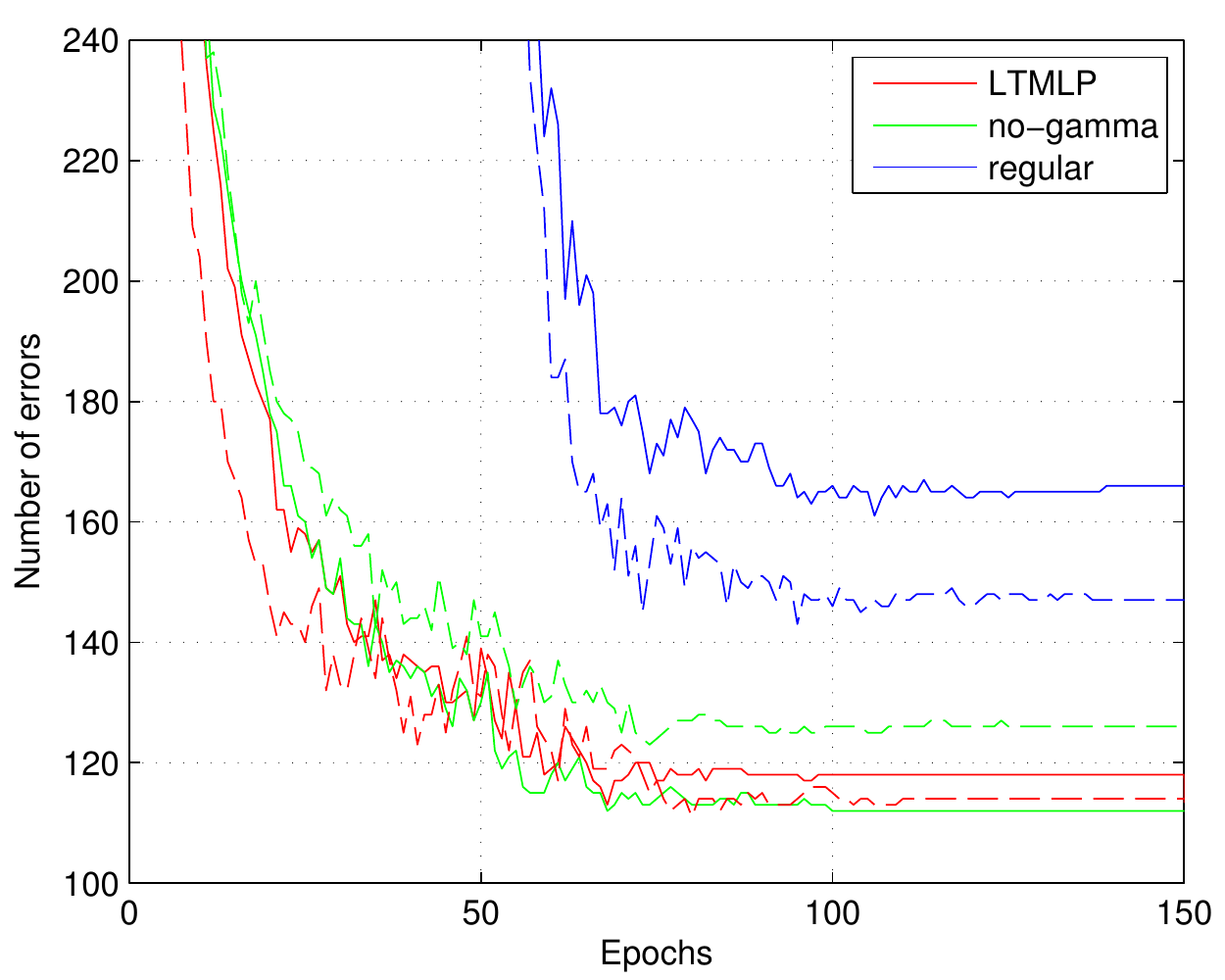}
\caption{The error rate on the MNIST test set for LTMLP training, LTMLP
without $\gamma$ and regular back-propagation. The solid lines show results for
networks with two hidden layers of 800 neurons and the dashed lines for
networks with three hidden layers of 400 neurons.}
\label{fig:mnist}
\end{figure}

\section{Experiments: MNIST Autoencoder}

\begin{figure}[tb]
\centering
\subfigure[MNIST classification]{
\includegraphics[height=3.7cm]{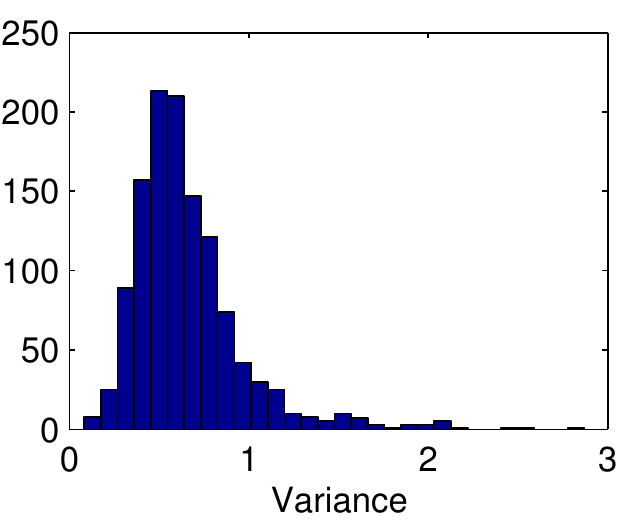}
}~
\subfigure[MNIST auto-encoder]{
\includegraphics[height=3.7cm]{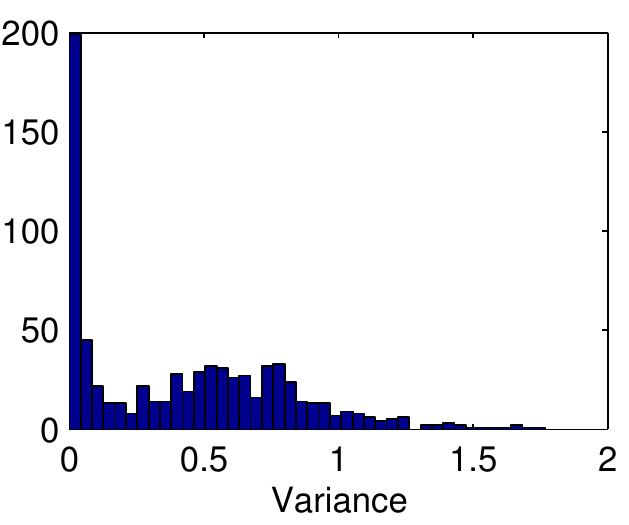}
}~
\subfigure[MNIST auto-encoder]{
\includegraphics[height=3.7cm]{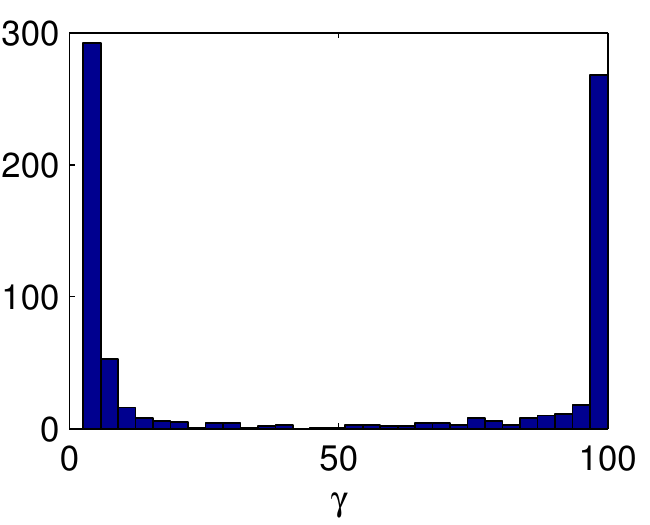}
}
\caption{Histograms of (a-b) variation of output of hidden neurons given the
MNIST training data and (c) $\gamma$s of the decoder part (4th and 5th hidden
layer) in the MNIST auto-encoder. (a) shows a healthy distributions of
variances, whereas in (b), which includes only variances of the decoder part,
there are many ``dead neurons''. These neurons induce corresponding $\gamma$s,
histogram of which is shown in (c), to blow up which eventually lead to
divergence.}
\label{fig:gammahist}
\end{figure}

Previously, we have studied an auto-encoder network using two transformations, $\alpha$ and
$\beta$, in \cite{raiko2012}. Now we use the same auto-encoder
architecture, 784--500--250--30--250--500--784. Adding the third
transformation $\gamma$ for training the auto-encoder poses problems.
Many hidden neurons in decoding layers (i.e., 4th and 5th hidden
layers) tend to be relatively inactive in the beginning of training, which
induces corresponding $\gamma$s to obtain very large values. In our
experiments, auto-encoder with $\gamma$s eventually diverge despite simple 
constraint we experimented with, such as $\gamma_i \leq 100$. This behavior is
illustrated in Figure~\ref{fig:gammahist}. The subfigure (a) shows
the distribution of variances of outputs of all hidden neurons in MNIST
classification network used in Section~\ref{sec:classification} given the MNIST
training data. The corresponding distribution for hidden neurons in
the decoder part of the auto-encoder is shown in the subfigure (b). The ``dead
neurons`` can be seen as a peak in the origin. The corresponding $\gamma$s,
constrained $\gamma_i \leq 100$, can be seen in the subfigure (c). We hypothesize
that this behavior is due to the fact, that in the beginning of the
learning there is not much information reaching the bottleneck layer
through the encoder part and thus there is nothing to learn for the decoding
neurons.  According to our tentative experiments, the problem described above
may be overcome by disabling $\gamma$s in the decoder network (i.e.,
fix $\gamma = 1$). However, this does not seem to speed up the learning 
compared to our earlier results with only two transformations in
\cite{raiko2012}. It is also be possible to experiment with weight-sharing or
other constraints to overcome the difficulties with $\gamma$s.

%






%

%
%
\section{Discussion and Conclusions}

We have shown that introducing linear transformation in nonlinearities
significantly improves the back-propagation learning in (deep) MLP networks.
In addition to two transformation proposed earlier in \cite{raiko2012}, we
propose adding a third transformation in order to push the Fisher information
matrix closer to unit matrix (apart from its scale). The hypothesis proposed in
\cite{raiko2012}, that the transformations actually mimic a second-order update
rule, was confirmed by experiments comparing the networks with transformations
and regular MLP network to a second-order update method. However, in order to
find out whether the third transformation, $\gamma$, we proposed in this paper,
is really useful, more experiments ought to be conducted. It might be useful to
design experiments where convergence is usually very slow, thus revealing
possible differences between the methods. As hyperparameter selection and
regularization are usually nuisance in practical use of neural networks, it
would be interesting to see whether combining dropouts \cite{hinton2012} and our
transformations can provide a robust framework enabling training of robust
neural networks in reasonable time.



The effect of the first two transformations is very similar to gradient factor
centering \cite{Schraudolph98factors,Schraudolph98decomposition}, but
transforming the model instead of the gradient makes it
easier to generalize to other contexts: When learning by by MCMC, variational
Bayes, or by genetic algorithms, one would not compute the basic
gradient at all. For instance, consider using the Metropolis algorithm
on the weight matrices, and expecially matrices $\mA$ and $\mB$. Without
transformations, the proposed jumps would affect the expected output $\vy_t$ and
the expected linear dependency $\partial \vy_t / \partial \vx_t$ in Eqs.\
\eqref{eq:expected_y}--\eqref{eq:expected_dydx}, thus often leading to low
acceptance probability and poor mixing. With the proposed transformations
included, longer proposed jumps in $\mA$ and $\mB$ could be accepted, thus
mixing the nonlinear part of the mapping faster. For further discussion, see
\cite{raiko2012}, Section 6. The implications of the
proposed transformations in these other contexts are left as future work.

\bibliographystyle{plain}
\bibliography{iclr2013}

\section*{Appendix}

\subsection*{Details of Section~\ref{sec:hessian}}

In experiments of Section~\ref{sec:hessian}, networks with all three
transformations (LTMLP), only $\alpha$ and $\beta$
(no-gamma) and network with no transformations (regular) were compared. 
Full batch training without momentum was used to make things as simple as
possible. The networks were regularized using weight decay and
adding Gaussian noise to the training data. Three hyperparameters, weight decay
term, input noise variance and learning rate, were validated for all networks
separately. The input data was normalized to zero mean and the network was
initialized as proposed in \cite{Glorot10}, that is, the weights were drawn from
a uniform distribution between $\pm \sqrt{6} / \sqrt{n_j+n_{j+1}}$, where $n_j$
is the number of neurons on the $j$th layer.

We sampled the three hyperparameters randomly (given our best guess intervals)
for 500 runs and selected the median of the runs that resulted in the best 50
validation errors as the hyperparameters. Resulting hyperparameters are
listed in Table~\ref{table:hyperparams}. Notable differences occur in step
sizes, as it seems that networks with transformations allow using significantly
larger step size which in turn results in more complete search in the weight
space.

\begin{figure}[tb]
\centering
\subfigure[Training error]{
\includegraphics[width=5.9cm]{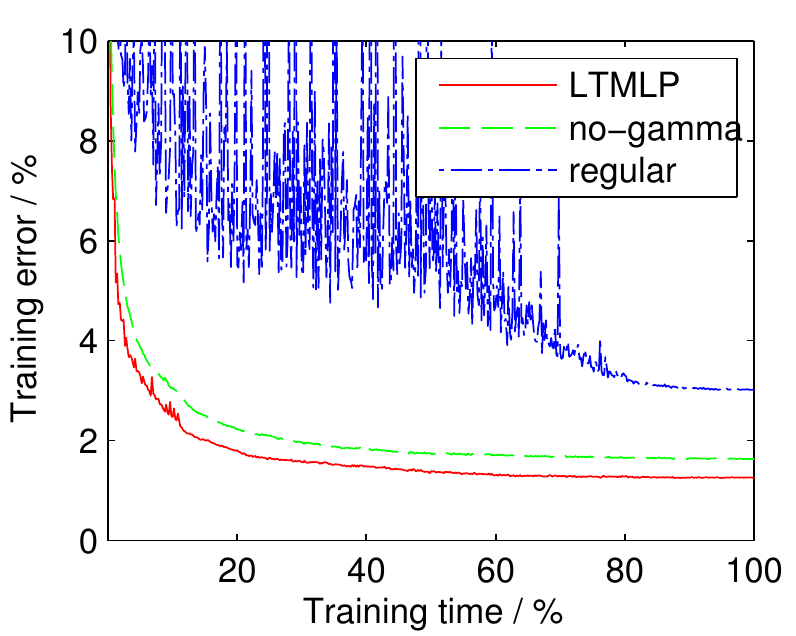}
}~
\subfigure[Test error]{
\includegraphics[width=5.9cm]{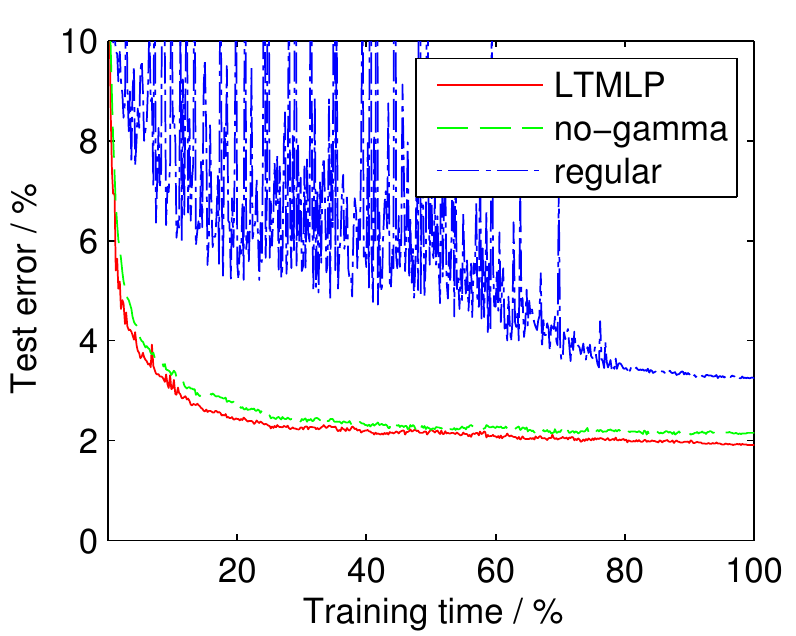}
}
\caption{Comparison of (a) training and (b) test errors of the algorithms
using the MNIST data in the experiment comparing them to the second-order
method. Note how the best learning for the regular MLP is relatively high, leading to oscillations until it is annealed towards the end.}
\label{fig:mnist_errors}
\end{figure}

Our weight updates are given by
\begin{align}
 \TT^\tau &= \TT^{\tau-1} - \varepsilon^\tau \nabla \TT^\tau.
\end{align}
where the learning rate on iteration $\tau$, $\varepsilon^\tau$, is given by 
\begin{align}
  \varepsilon^\tau &= \left\{\begin{array}{ll} \varepsilon_0 & \tau \leq T/2
\\   2(1-\frac{\tau}{T})\varepsilon_0 & \tau > T/2 \end{array} \right.
\end{align}
that is, the learning rate starts decreasing linearly after the midpoint of the
given training time $T$. Furthermore, the learning rate $\varepsilon^\tau$ is
dampened for shortcut connection weights by multiplying with
$\left(\frac{1}{2}\right)^s$, where $s$ is number of skipped
layers as proposed in \cite{raiko2012}.\footnote{This heuristic is not well supported by analysis of Figure \ref{fig:diagvalues} and could be re-examined.}
Figure~\ref{fig:mnist_errors} shows training and test errors for the
networks. The LTMLP obtains the best results although there is no big difference
compared to training without $\gamma$.

\begin{table}[b]
\caption{Hyperparameters for the neural networks}
\vspace{4mm}
\label{table:hyperparams}
\begin{center}
\begin{tabular}{c|ccc}
& LTMLP & no-gamma & regular\\
\hline
weight decay& $4.6\times10^{-5}$ & $1.3\times10^{-5}$ & $3.9\times10^{-5}$\\
noise & 0.31 & 0.36 & 0.29\\
step size & 1.2 & 2.5 & 0.45\\
\end{tabular}
\end{center}
\end{table}

\subsection*{Details of Section~\ref{sec:classification}}

The MNIST dataset consists of $28\times28$ images of hand-drawn digits. There
are 60\ 000 training samples and 10\ 000 test samples. We experimented with
two networks with two and three hidden layers and number of hidden neurons by
arbitrary choice. Training was done in minibatch mode with 1000 samples in each
batch and transformations are updated on every iteration using the current
minibatch with using \eqref{eq:alpha}--\eqref{eq:gamma}. This seems to
speed up learning compared to the approach in \cite{raiko2012} where
transformations were updated only occasionally with the full training data.
Random Gaussian noise with $\sigma=0.3$ was injected to the training data in the
beginning of each epoch.

Our weight update equations are given by:
\begin{align}
 \Delta \TT^\tau &= \nabla \TT + p^\tau \Delta \TT^{t-1}, \\
 \TT^\tau &= \TT^{\tau-1} - \varepsilon^\tau \Delta \TT^\tau,
\end{align}
where
\begin{align}
\varepsilon^\tau &= \left\{\begin{array}{ll} \varepsilon_0 &  \tau \leq T \\
\varepsilon_0 f^{\tau-T} & \tau > T \end{array} \right. \\
p^\tau &= \left\{ \begin{array}{ll} \frac{\tau}{T} p_f + (1- \frac{\tau}{T})p_0
& \tau
\leq T \\
p_f & \tau > T \end{array} \right.
\end{align}
In the equations above, $T$ is a ``burn-in time'' where momentum $p^\tau$ is
increased from starting value $p_0 = 0.5$ to $p_f = 0.9$ and learning rate
$\varepsilon = \varepsilon_0$ is kept constant. When $\tau>T$ momentum is kept
constant and learning rate starts decreasing exponentially with $f=0.9$.
Hyperparameters were not validated but chosen by arbitrary guess such that
learning did not diverge. For the regular training, $\varepsilon_0 = 0.05$ was
selected since it diverged with higher learning rates. Then according to lessons
learned, e.g. in Section \ref{sec:hessian}, $\varepsilon_0 = 0.3$ was set for
LTMLP with $\gamma$ and $\varepsilon_0 = 0.7$ for the variant with no $\gamma$.
Basically, it seems that transformations allow using higher learning rates
and thus enable faster convergence.

\end{document}